\documentclass{article}
\usepackage[utf8]{inputenc}
\makeatletter
\providecommand\input@path {}
\def\addInputPath#1{\xdef\input@path{\unexpanded\expandafter{\input@path}{\unexpanded{#1}}}}

\addInputPath{NIPS/}
\makeatother
\usepackage[final,nonatbib]{nips_2018}
\usepackage[numbers,square]{natbib}
\usepackage[T1]{fontenc}    
\usepackage{url}            
\usepackage{booktabs}       
\usepackage{amsfonts}       
\usepackage{nicefrac}       
\usepackage[
protrusion=alltext-nott, 
final]{microtype}      

\usepackage{amsmath}
\usepackage[inline]{enumitem}

\usepackage{graphicx} 

\usepackage{subcaption}
\input{fig/rddc/tikzhelpers}
\newcommand\RNN[1][]{
\begin{scope}[#1]
\tikzNN[
  name=NNinRNN,
	nn neurons={4,4,4},
	inside layer style={fill=blue!40},
	layer styles={{fill=red!80},{fill=green!80},{fill=magenta!80}},
	scale={1},
	layer padding={0.05cm},
	layer height={1.5cm},
	layer sep={.5cm},
	neuron radius={0.1cm},
	connection style={gray},
	neuron color={gray!20},
	neuron style={},
  tikz args={}
]
\xdef\sqpadding{0.04}
\begin{pgfonlayer}{wallpaper}
  \draw[fill=black!30] ($(NNinRNN-bb.south west)+(-\sqpadding,-\sqpadding)$) rectangle node[fitting node](RNN-bb){} ($(NNinRNN-bb.north east)+(\sqpadding,\sqpadding)$);
\end{pgfonlayer}
\node[below] at (RNN-bb.south){RNN};
\xdef\arrdist{0.5}
\draw[->, thick] ($(RNN-bb.north)+(\arrdist,0)$) arc(-20:200:\arrdist);
\end{scope}%
}

\makeatletter
\newif\ifNN@hideLayerBox

\pgfkeys{
	/NN/.cd,
	connection style/.meta tikz style={NN/connection style},
	neuron style/.meta append tikz style={NN/neuron style}{fill=\nn@cfg{neuron color}},
	inside layer style/.meta tikz style={NN/inside layer style},
	hide layer boxes/.is if={NN@hideLayerBox},
	user defaults/.settable style,
	user defaults={}
}
\pgfkeys{
	/NN/factory defaults/.style={
			/NN/.cd,
      name/.initial={NN},
			input node name/.initial={NN.in},
			output node name/.initial={NN.out},
			nn neurons/.initial={4,2,1},
			layer styles/.initial={,},
			scale/.initial={1},
			layer padding/.initial={0.05cm},
			layer height/.initial={3cm},
			layer sep/.initial={1cm},
			neuron radius/.initial={0.2cm},
			connection style={gray},
			neuron color/.initial={gray!50},
			neuron style={},
			inside layer style={fill=red!40},
      tikz args/.initial={}
		}
}

\def\nn@cfg#1{\csname pgfk@/NN/#1\endcsname}
\let\nn@c\nn@cfg
\newif\ifisfirstLayer
\newdimen\firstneuron@x
\newcommand\tikzNN[1][]{%
	\pgfexpkeys{/NN/.cd,/NN/factory defaults,/NN/user defaults,#1}
  \xdef\@tNN{\nn@c{name}}
	\edef\scaleArgs{\nn@c{tikz args},scale=\nn@c{scale}, every node/.style={scale=\nn@c{scale}}}
	\edef\scopeArgs{{scope}[\scaleArgs]}
	\expandafter\begin\scopeArgs

	\def\layerIdx{0}
	\edef\neuronlist{\nn@c{nn neurons}}
  \readlist*\neurItemList\neuronlist
	\edef\@styleitms{\nn@c{layer styles}}
	\readlist\layerStylesList\@styleitms
  \providecommand\@fitnodesList{}
  \newcommand\addfitnode[1]{\xdef\@fitnodesList{\unexpanded\expandafter{\@fitnodesList}(##1)}}

	\foreach\noNeurons in \neuronlist{
		\edef\layerIdxOne{\the\numexpr\layerIdx+1}
		\pgfmathsetlengthmacro\neurX{\nn@c{layer sep}*\layerIdx}
		\pgfmathsetlengthmacro\neuronsep{\nn@c{layer height}/\noNeurons}
		\pgfkeys{/tikz/NN/this layer style/.style={}}
		\ifnum\layerIdx<\layerStylesListlen\relax
			\edef\thislayerstyleargs{{/tikz/NN/this layer style/.style=\layerStylesList[\layerIdxOne]}}
			\expandafter\pgfkeys\thislayerstyleargs
		\fi
		\ifNN@hideLayerBox\else
			\draw[NN/inside layer style, NN/this layer style]%
			($(\neurX-\nn@c{neuron radius}-\nn@c{layer padding}, 0)$) rectangle node[fitting node](\@tNN-rectangle-\layerIdx){} ++($(2*\nn@c{neuron radius}+2*\nn@c{layer padding},\neuronsep*\noNeurons)$);
      \addfitnode{\@tNN-rectangle-\layerIdx}
		\fi
		\foreach\neurno in {1,...,\noNeurons}{
				\edef\neurIdx{\the\numexpr\neurno-1}
				\pgfmathsetlengthmacro\neurY{\neuronsep*\neurIdx+\neuronsep/2}
				\node(\@tNN-neur-\layerIdx-\neurIdx) at (\neurX,\neurY){};
				\draw[NN/neuron style,fill=\nn@cfg{neuron color}](\@tNN-neur-\layerIdx-\neurIdx) circle (\nn@c{neuron radius});
        \addfitnode{\@tNN-neur-\layerIdx-\neurIdx}
			}
		\xdef\layerIdx{\the\numexpr\layerIdx+1}
	}
  \edef\lastLayer{\the\numexpr\layerIdx-1}
  \edef\lastNeuron{\the\numexpr\neurItemList[-1]-1}
	\edef\numLayers{\the\numexpr\layerIdx+1}
	\edef\numSourceLayers{\the\numexpr\layerIdx-2}
	\begin{pgfonlayer}{background}
		\readlist\layersdict\neuronlist
		\foreach\srcLayer in {0,...,\numSourceLayers}{
				\edef\dstLayer{\the\numexpr\srcLayer+1}
				\edef\numNeuronsInSrcLayer{\layersdict[\the\numexpr\srcLayer+1]}
				\edef\numNeuronsInDstLayer{\layersdict[\the\numexpr\srcLayer+2]}
				\foreach\srcNeuron in {0,...,\the\numexpr\numNeuronsInSrcLayer-1}{
						\foreach\dstNeuron in {0,...,\the\numexpr\numNeuronsInDstLayer-1}{
								\draw[NN/connection style](\@tNN-neur-\srcLayer-\srcNeuron) -- (\@tNN-neur-\dstLayer-\dstNeuron);
							}
					}
			}
	\end{pgfonlayer}
  \pgfmathsetlengthmacro\Lnnwid{(2*\nn@c{neuron radius}+2*\nn@c{layer padding})+(\neurItemListlen-1)*\nn@c{layer sep}}
  \pgfmathsetlengthmacro\Lnnheight{\nn@c{layer height}}
  \xdef\nnwid{\the\dimexpr\Lnnwid-\nn@c{neuron radius}-\nn@c{layer padding}}
  \xdef\nnheight{\Lnnheight}
  \pgfextractx{\firstneuron@x}{\pgfpointanchor{\@tNN-neur-0-0}{center}}%
  \xdef\nnOriginX{\the\dimexpr\firstneuron@x-\nn@c{neuron radius}-\nn@c{layer padding}}
  \xdef\nnOrigin{\nnOriginX,0}
  \coordinate(\@tNN-origin) at (\nnOrigin);
  \coordinate(\@tNN-topRight) at (\nnwid,\nnheight);
	\end{scope}
  \draw[fill=none,draw=none] (\@tNN-origin) rectangle node[fitting node](\@tNN-bb){} (\@tNN-topRight);
}

\makeatletter\if@preprint
\def\todoopt{[]}
\else
\def\todoopt{[disable]}
\fi
\expandafter\usepackage\todoopt{todonotes}

\input{commands}

\newcommand\blfootnote[1]{%
  \begingroup
  \renewcommand\thefootnote{}\footnote{#1}%
  \addtocounter{footnote}{-1}%
  \endgroup
}

\title{Recurrent Deep Divergence-based Clustering for simultaneous feature learning and clustering of variable length time series}

\def\mlgroup{UiT Machine-Learning Group}
\def\uit{UiT The Arctic University of Norway}
\author{%
	Daniel J.~Trosten \\
  \mlgroup \\
  \uit \\
	\texttt{daniel.j.trosten@uit.no} \\
	\And
	Andreas S.~Strauman \\
  \mlgroup \\
  \uit \\
  \And
  Michael Kampffmeyer \\
  \mlgroup \\
  \uit \\
  \And
  Robert Jenssen \\
  \mlgroup \\
  \uit \\
}
\date{November 2018}

\begin{document}
  \maketitle

  \blfootnote{This work was partially funded by the Norwegian Research Council FRIPRO grant no. 239844.}
  \blfootnote{Copyright 2019 IEEE. Published in the IEEE 2019 International Conference on Acoustics, Speech, and Signal Processing (ICASSP 2019), scheduled for 12-17 May, 2019, in Brighton, United Kingdom. Personal use of this material is permitted. However, permission to reprint/republish this material for advertising or promotional purposes or for creating new collective works for resale or redistribution to servers or lists, or to reuse any copyrighted component of this work in other works, must be obtained from the IEEE. Contact: Manager, Copyrights and Permissions / IEEE Service Center / 445 Hoes Lane / P.O. Box 1331 / Piscataway, NJ 08855-1331, USA. Telephone: + Intl. 908-562-3966.}

  \begin{abstract}
    The task of clustering unlabeled time series and sequences entails a particular set of challenges, namely to adequately model temporal relations and variable sequence lengths. If these challenges are not properly handled, the resulting clusters might be of suboptimal quality.
    As a key solution, we present a joint clustering and feature learning framework for time series based on deep learning. For a given set of time series, we train a recurrent network to represent, or embed, each time series in a vector space such that a divergence-based clustering loss function can discover the underlying cluster structure in an end-to-end manner.
    Unlike previous approaches, our model inherently handles multivariate time series of variable lengths and does not require specification of a distance-measure in the input space.
    On a diverse set of benchmark datasets we illustrate that our proposed Recurrent Deep Divergence-based Clustering approach outperforms, or performs comparable to, previous approaches.
  \end{abstract}

  \section{Introduction}

The vast amounts of complex data that need to be categorized in an unsupervised manner, makes clustering \cite{theodoridis_pattern_2009,shalev-shwartz_understanding_2014} one of the key areas in machine learning and of growing importance. In many cases it is unrealistic, or even infeasible, to label individual data points for supervised learning.

The majority of classical clustering algorithms requires the data to reside in a vector space equipped with some distance function or similarity measure. However, for complex datatypes, such as images or sequences, this requirement is not necessarily met. Much research in the machine learning field has therefore been invested in the development of feature extraction techniques for such datatypes. These produce vectorial representations embedded in a space with a suitable distance measure. Such methods are often computationally complicated procedures that may not be robust across different domains and data types. Post computation, the features can be clustered using e.g. \( k \)-means \cite{macqueen_methods_1967}, Hierarchical Clustering \cite{jr_hierarchical_1963}, or Spectral Clustering \cite{shi_normalized_2000}. However, there is no guarantee that the extracted features are well suited for the selected clustering algorithm, which causes the quality of the resulting clusters to depend heavily on the representation.

Supervised deep learning has seen tremendous recent developments for end-to-end representation learning \cite{bengio_representation_2013}, wherein the data representation is obtained as an integral part of the optimization of the neural network classifier \cite{krizhevsky_imagenet_2012,cho_learning_2014}. The translation of these achievements to the unsupervised case of clustering, has been hailed as a main next goal in machine learning \cite{lecun_deep_2015}. Several works have been proposed along these lines over the last couple of years, nevertheless such research is still in its infancy.

Deep Embedded Clustering (DEC \cite{xie_unsupervised_2016} and IDEC \cite{guo_improved_2017}),  the Deep Clustering Network (DCN) \cite{yang_towards_2017}, and the Categorical GAN (CatGAN) \cite{springenberg_unsupervised_2015}, are some examples of novel unsupervised deep learning architectures. In these models the raw input signal is processed by a deep neural network, producing a vectorial representation. Based on this representation, the subsequent parts of the model then computes the cluster membership prediction. In DEC, for example, a set of inputs are processed by a Multilayer Perceptron (MLP) to produce a corresponding set of hidden representations.
The hidden representations are then softly assigned to a set of centroids, based on Euclidean distance in the space of hidden representations. The joint optimization of MLP-parameters and centroids then allows the feature generating MLP to adapt based on the clustering of the hidden representations.
The MLP is pre-trained as a stacked 
autoencoder to ensure that the hidden representations preserve some of the structure present in the input space.

Another recent architecture that incorporates similar ideas, is Deep Divergence-based Clustering (DDC) \cite{kampffmeyer_deep_2017-1}. DDC was originally designed for image clustering, and therefore uses a convolutional neural network (CNN) for feature extraction. Cluster assignments are obtained by a clustering module based on information theoretic quantities computed using the representations produced by the CNN. Moreover, DDC does not require autoencoder initialization, and can therefore be trained from start to finish without modifications to the architecture.

On the other hand, when it comes to the virtually omnipresent domain of sequential data, none of the aforementioned end-to-end clustering methods are directly applicable. Learning to Cluster (L2C) \cite{meier_learning_2018} is a model designed for deep learning-based sequence clustering, but requires pairwise weakly labeled observations during training, and is therefore not fully unsupervised.

In this paper we propose a novel end-to-end architecture for joint representation learning and clustering of sequential data. Our model aims to address some of the challenges that arise when modeling sequential data, namely variable sequence length, multivariate elements, and complex temporal dependencies. We do this by integrating a recurrent neural network within an architecture building on the DDC framework, which we refer to as Recurrent Deep Divergence-based Clustering (RDDC).\par By this, we leverage the power of DDC which has proven to perform well on image clustering without relying on extra model components for initialization.

The rest of the paper is structured as follows; Section 2 introduces the different components of our model and how they interact. In Section 3 we provide some experimental results, as well as a qualitative analysis for one of the experimental cases. Finally, we make some concluding remarks in Section 4.

  \section{Method}

\bgroup
\makeatletter
\def\input@path{{fig/rddc/}}
\begin{figure}[t]
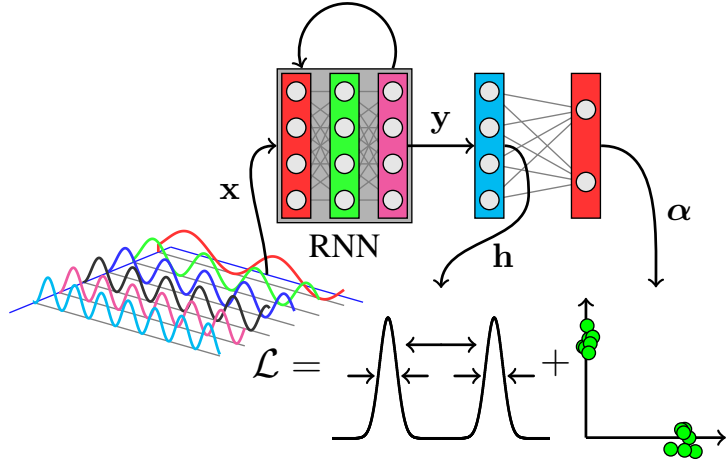

  \centering
  \resizebox{0.7\textwidth}{!}{
\makeatletter
\tikzset{connection/.style={->,thick}}
\makeatletter
\begin{tikzpicture}
\RNN[shift={(0,0)}]
\tikzNN[
  name={MAIN-NN},
	nn neurons={4,2},
  neuron radius={0.1cm},
	inside layer style={fill=blue!40},
  layer styles={{fill=cyan!80},{fill=red!80}},
	layer height={1.5cm},
  neuron color={gray!20},
  tikz args={shift={(2,0)}}
]
\begin{scope}[shift={(-1.3cm,-0.3cm)},scale=0.4]
\input{timeseries}
\end{scope}
\draw[connection] (times-anchor) to[out=90, in=180] node[left]{$\vec x$} (RNN-bb.west);
\extractCoords[south]{MAIN-NN-bb}{mainbb@s}
\def\mainbbDist{1.5cm}
\edef\xdist{1cm}
\def\sepcompwd{2.25cm}
\def\sepcompht{1.55cm}

\begin{scope}[shift={(\the\dimexpr\mainbb@s@x-\xdist,\the\dimexpr\mainbb@s@y-\sepcompht/2-\mainbbDist)}]
  \input{sepcomp.tikz}
  \draw[draw=none, fill=none](\the\dimexpr-\sepcompwd/2,0) rectangle node[fitting node](sepcomp-bb){} (\sepcompwd/2,\sepcompht);
\end{scope}
\extractCoords[north]{sepcomp-bb}{sepcomp@n}
\extractCoords[south]{sepcomp-bb}{sepcomp@s}
\edef\leftoffset{0.5cm}

\begin{scope}[shift={(\the\dimexpr\mainbb@s@x-\leftoffset+\xdist,\sepcomp@n@y-\sepcompht)}]
  \tikzset{point/.style={circle,fill=green,inner sep=0,minimum size=4pt, draw=black}}
  \edef\arrlen{\the\dimexpr\sepcompht-0.1cm}
  \draw[draw=none](0,0) rectangle node[fitting node](simplex-bb){} (\arrlen,\sepcompht);
  \draw[->, thick](0,0) --++ (0,\arrlen);
  \draw[->, thick](0,0) --++ (\arrlen,0);
  \input{simplexPoints}
\end{scope}
\draw[connection] (NNinRNN-bb.east) -- node[above]{$\vec y$} (MAIN-NN-bb.west);
\draw[connection] (MAIN-NN-bb.east) to [out=0, in=90] node[right,pos=0.6]{$\gvec\alpha$} (simplex-bb.north);
\draw[connection] (MAIN-NN-rectangle-0.east) to[out=0,in=90] ++(0.25cm,-0.5cm) to[out=-90, in=90] node[pos=0.7,right=0.2cm]{$\vec h$} (sepcomp-bb.north);
\node at ($(sepcomp-bb.west)+(-0.1cm,0)$){\llap{\large$\mathcal{L}=$}};
\coordinate (tmp) at ($(simplex-bb.west)-(\leftoffset/2,0)$);
\node at ($(sepcomp-bb.east)!0.5!(tmp)$){\large$\mathbf+$};
\end{tikzpicture}}
  \caption{An overview of the RDDC architecture. When the input is processed by the RNN, two fully connected layers extract the learned feature \( \vec h \), and cluster assignment \( \gvec\alpha \), respectively.}
  \label{fig:ddc}
\end{figure}
\egroup

A conceptual overview of the RDDC architecture is provided in Fig. \ref{fig:ddc}. Suppose we have \( n \) input sequences \( \vec x_1, \dots \vec x_n \) to cluster. First, these are processed by the RNN, which is two-layer bidirectional Gated Recurrent Unit \cite{cho_learning_2014}.
The final hidden states of the RNN are concatenated and passed on to a Batch Normalization transformation \cite{ioffe_batch_2015} producing the intermediate variables \( \vec y_1, \dots, \vec y_n \). Subsequently, they are transformed by the first fully connected layer to obtain the hidden representations \( \vec h_1, \dots \vec h_n \). Finally, the hidden representations are passed through the fully connected output layer with a softmax activation function, to produce the (soft) cluster membership vectors \( \gvec \alpha_1, \dots, \gvec \alpha_n \).

\subsection{Loss function}
	The model is trained end-to-end using a loss function which is designed with three key properties in mind:
	\begin{enumerate*}[label=(\roman*)]
		\item Cluster separability and compactness.
    \item Cluster orthogonality in the observation space.
		\item Closeness of cluster memberships to a simplex corner.
	\end{enumerate*}

	The DDC loss function consists of three terms. The first term tackles the separability and compactness property outlined above.
   Consider the multiple-pdf generalization of the Cauchy-Schwartz (CS) divergence \cite{jenssen_cauchyschwarz_2006}:
	\begin{align}
		\label{eq:D_cs}
		D_{cs} = - \log\lrp{ \frac{1}{k} \sums{i=1}{k-1} \sums{j>i}{} \frac{\E_{\vec H \sim p_i}(p_j(\vec H))}{\sqrt{ \E_{\vec H \sim p_i}(p_i(\vec H)) \E_{\vec H \sim p_j}(p_j(\vec H))}} }
	\end{align}
	where \( k \) is the number of distributions, and \( \E_{\vec X \sim p}(g(\vec X)) \) denotes the expectation of \( g(\vec X) \) when \( \vec X \) has distribution \( p \).
	If we let each \( p_i \) represent a cluster, a large divergence would lead to well separated and compact clusters. Maximizing \eqref{eq:D_cs} is equivalent to minimizing the argument of the logarithm, which gives the loss term
	\begin{align}
		\label{eq:L1_ints}
		\cl L_1 = \frac{1}{k} \sums{i=1}{k-1} \sums{j>i}{} \frac{\E_{\vec H \sim p_i}(p_j(\vec H))}{\sqrt{ \E_{\vec H \sim p_i}(p_i(\vec H)) \E_{\vec H \sim p_j}(p_j(\vec H))}}.
	\end{align}
  Using the kernel density estimator \cite{parzen_estimation_1962} with a Gaussian kernel to estimate \( p_1, \dots, p_k \) gives
  \begin{align}
    \label{eq:L1}
    \cl L_1 = \frac{1}{k} \sums{i=1}{k-1}\sum_{j > i} \frac{\v a_i\T \mat K_h \v a_j}{\sqrt{\v a_i\T \mat K_h \v a_i \v a_j\T \mat K_h \v a_j}}
  \end{align}
  where \( \mat K_h = [k_{lm}] \), \( k_{lm} = \exp\lrp{-\frac{||\v h_l - \v h_m||^2}{2\sigma^2}} \). The vectors \( \v a_1, \dots \v a_k \) denote the \emph{columns} of the \( n \times k \) hard cluster assignment matrix \( \mat A \). During optimization, we relax the hard membership constraint to make the loss-function differentiable. Thus, we can form \( \mat A \) by stacking the soft cluster assignment vectors \( \gvec\alpha_1, \dots, \gvec\alpha_n \) row-wise.

  The second term in the loss function is designed such that the clusters are orthogonal in the \( n \)-dimensional observation-space. This accounts to the matrix \( \mat A \) having orthogonal columns, which motivates the loss
  \[ \cl L_2 = \triu(\mat A\T \mat A), \]
  where \( \triu(\cdot) \) denotes the sum of the strictly upper triangular elements of its argument. Note that this term differs from the term \( \triu(\mat A \mat A\T) \), which was originally used for DDC~\cite{kampffmeyer_deep_2017-1}. In contrast to the original loss term, our formulation does not introduce a regularizing effect to the computation of cluster membership vectors. In our experiments, we found that model performance improved when said regularization effect was removed.


	The final term in the loss function is constructed such that the cluster membership vectors lie close to a corner of the simplex defined by the softmax activation function. Let \( \mat M =  [m_{li}],~ m_{li} = \exp(-||\gvec \alpha_l - \vec e_i||^2) \), where \( \vec e_i \) is the \( i \)-th cartesian basis vector of \( \real^k \) (\( i \)-th corner of the simplex). Then, the last loss term is
	\begin{align}
		\label{eq:L3}
		\cl L_3 = \frac{1}{k} \sums{i=1}{k-1}\sum_{j > i} \frac{\v m_i\T \mat K_h \v m_j}{\sqrt{\v m_i\T \mat K_h \v m_i \v m_j\T \mat K_h \v m_j}}
	\end{align}
	which is very similar to \eqref{eq:L1}, but \( \v a_1, \dots, \v a_k \) have now been replaced with \( \v m_1, \dots, \v m_k \), which denote the columns of \( \mat M \). If we consider \( m_{qi} \) as the soft assignment of cluster membership vector \( q \) to simplex corner \( i \), we can interpret \( \cl L_3 \) in the same way as \( \cl L_1 \): The distributions of cluster assignment vectors should be compactly centered around distinct simplex corners.

	The total loss function is a linear combination of the three loss terms:
  \[ \cl L = \cl L_1 + w_2 \cl L_2 + w_3 \cl L_3 \]
  where \( w_2 \) and \( w_3 \) are hyperparameters.

  \section{Experiments}

\subsection{Experiment setup}
To quantitatively evaluate the performance of our model on sequential data, we use four different datasets. These were selected as each of them represents a distinct, but commonly observed, sequence-generating process. The variation in sequence length and dimensionality across the datasets should provide broad insight into the capabilities of the model. The datasets are:
\begin{itemize}
  \item \emph{Character Trajectories} (CT) \cite{dheeru_uci_2017}. The sequences form trajectories of handwritten characters. A subset consisting of the characters \( \{a,b,c,d,e,g \} \) was chosen for evaluation.

  \item \emph{Twenty Newsgroups} (TN) \cite{dheeru_uci_2017,cardoso-cachopo_improving_2007}. This dataset contains news articles from different categories. Following the example of \cite{yu_document_2010,kim_thematic_2012} we choose a subset of the data containing articles from distinct domains. The subset consists of articles from \texttt{alt.atheism}, \texttt{comp.graphics} and \texttt{misc.forsale}, with lengths between \( 50 \) and \( 300 \) words.
  The articles were converted to \( 100 \)-dimensional sequences using a Skipgram-Word2Vec model \cite{mikolov_efficient_2013}.

  \item \emph{Speech Commands} (SC) \cite{warden_speech_2017}: Each time series is a raw sound recording of a single spoken English word. For evaluation, a subset consisting of the words \( \{ \text{Yes, No} \} \) were used.
  {\setlength\parskip{0pt}\par Prior to being analyzed by the network, the data was preprocessed in the following manner:}
  \begin{enumerate*}[label=(\roman*)]
    \item Crop to remove leading and trailing periods of low activity in the recording.
    \item Normalize such that each recording has zero mean and unit variance.
    \item Compute short-time log-frequency filter banks \cite{theodoridis_pattern_2009}, using \( 12 \) bins, a window length of \( 15 \) ms, and a window overlap of \( 7 \) ms.
  \end{enumerate*}

  \item \emph{Arabic Digits} (AD) \cite{dheeru_uci_2017}: The sequences consists of mel-frequency cepstrum coefficients obtained from recordings of spoken Arabic digits. For this dataset, all ten digits were used.
\end{itemize}

The datasets were divided into training, validation and test sets, each set receiving \( 80\,\% \), \( 10\,\% \) and \( 10\,\% \) of the samples, respectively. The training set was used for training, the validation set for hyperparameter tuning/model selection, and the test set for performance evaluation.

The model we use for testing is a two-layer bidirectional GRU with 32 units in each layer, followed by the two fully connected layers. The first fully connected layer has \( 16 \) units for the Character Trajectories dataset and \( 32 \) units for the Speech Commands, Twenty Newsgroups and Arabic Digits datasets. In our experience, the model was not particularly sensitive to the number of RNN or fully connected units. The number of units in the output layer is the same as the number of clusters in the dataset.

We compare our model to the following benchmark methods, which represent both classical clustering approaches, as well as more recent deep learning-based clustering approaches:
\begin{enumerate*}[label=(\roman*)]
  \item \( k \)-means \cite{macqueen_methods_1967}.
  \item Ward-linkage Hierarchical Clustering \cite{jr_hierarchical_1963}.
  \item Spectral Clustering \cite{shi_normalized_2000}.
  \item DEC with the configuration specified by the authors \cite{xie_unsupervised_2016}.
  \item DDC with just the last two fully connected layers \cite{kampffmeyer_deep_2017-1}.
\end{enumerate*}
As these methods all require vectorial inputs of fixed length, they are implemented using each of the following vectorization methods:
\begin{itemize}
  \item Zero padding: Each time series is augmented with zero-vectors such that its length matches the longest length in the dataset.
  \item Cropping\char`:\ All timesteps recorded after the shortest sequence-length in the dataset are discarded.
  \item Time averaging: The vector average along the time axis is computed for each sequence.
\end{itemize}
Finally, for the zero-padded and cropped sequences, we concatenate the remaining observations for each timestep, producing one vector for each sequence.

Our model is implemented in TensorFlow and trained on stochastic mini-batches of size \( 200 \), using the Adam optimizer \cite{kingma_adam:_2014}. Each DDC/RDDC model was trained for \( 150 \) epochs from \( 20 \) different initializations on each data set. The model resulting in the lowest value of the loss function was then selected for further evaluation. The kernel width, \( \sigma \) was set to \( 15\,\% \) of the median pairwise distance between the hidden representations \( \vec h \), within each batch, following \cite{jenssen_kernel_2010}. The median was computed during each forward pass and treated as fixed during the backward passes.  After each training run, the unsupervised clustering accuracy on the test set was computed as \( {\fam0 ACC} = \max\limits_{\cl M}\frac{1}{n} \sum_{i=1}^{n} \delta(l_i - \cl M(c_i)) \)
where \( l_i \) and \( c_i \) is the true label and the predicted cluster label of the \( i \)-th sequence, respectively. The maximum runs over bijective cluster-to-class maps, effectively finding the "best" cluster-to-class assignment in terms of classification accuracy. We also compute the normalized mutual information, defined as \( {\fam0 NMI} = \frac{2\, I(l,c)}{H(l) + H(c)} \)
where \( I(l,c) \) is the mutual information between the predicted cluster assignments and the true labels, and \( H(\cdot) \) denotes the entropy of its argument.

\subsection{Quantitative results}
\begin{table}
  \centering
  \caption{Resulting accuracy ({\( \fam0 ACC \)}) and normalized mutual information ({\( \fam0 NMI \)}) for the different models and datasets. Note that in the published version of the Arabic Digits dataset, all the sequences are normalized to have zero-mean, meaning that the time averaging vectorization technique is not applicable.}
  \label{tab:results}
\bgroup
\let\B\textbf
\renewcommand{\arraystretch}{1.2}
\begin{tabular}{c|cc|cc|cc|cc}
  \thickhline
  \multicolumn{1}{c}{} & \multicolumn{2}{c}{CT} & \multicolumn{2}{c}{TN} & \multicolumn{2}{c}{SC} & \multicolumn{2}{c}{AD} \\ \thickhline
  Model           & {\( \fam0 ACC \)}     & {\( \fam0 NMI \)}     & {\( \fam0 ACC \)}      & {\( \fam0 NMI \)}      & {\( \fam0 ACC \)}      & {\( \fam0 NMI \)}      & {\( \fam0 ACC \)}      & {\( \fam0 NMI \)} \\ \hline
  \( k \)-m (Zero) & \B{1.0} & \B{1.0} & 0.38     & 0.01     & 0.51     & 0.0      & 0.71     & 0.6 \\
  \( k \)-m (Crop) & \B{1.0} & \B{1.0} & 0.56     & 0.35     & 0.54     & 0.0      & 0.51     & 0.48 \\
  \( k \)-m (Avg.) & 0.87    & 0.79    & \B{0.96} & \B{0.88} & 0.58     & 0.03     & --       & -- \\\hline
  HC (Zero)       & \B{1.0} & \B{1.0} & 0.42     & 0.07     & 0.5      & 0.0      & 0.78     & 0.75 \\
  HC (Crop)       & \B{1.0} & \B{1.0} & 0.78     & 0.45     & 0.5      & 0.0      & 0.52     & 0.56 \\
  HC (Avg.)       & 0.87    & 0.84    & 0.78     & 0.53     & 0.57     & 0.04     & --       & -- \\\hline
  SC (Zero)       & 0.40    & 0.33    & 0.3      & 0.0      & 0.50     & 0.03     & 0.66     & 0.61 \\
  SC (Crop)       & 0.41    & 0.32    & 0.36     & 0.04     & 0.51     & 0.01     & 0.47     & 0.45 \\
  SC (Avg.)       & 0.69    & 0.67    & 0.95     & 0.8      & 0.51     & 0.01     & --       & -- \\ \hline
  DEC (Zero)      & \B{1.0} & \B{1.0} & 0.37     & 0.01     & 0.51     & 0.0      & 0.66     & 0.67 \\
  DEC (Crop)      & \B{1.0} & \B{1.0} & 0.6      & 0.31     & 0.53     & 0.0      & 0.51     & 0.47 \\
  DEC (Avg.)      & 0.54    & 0.66    & \B{0.96} & 0.85     & 0.52     & 0.0      & --       & -- \\\hline
  DDC (Zero)      & 0.98    & 0.96    & 0.41     & 0.02     & 0.54     & 0.0      & 0.61     & 0.59 \\
  DDC (Crop)      & \B{1.0} & \B{1.0} & 0.49     & 0.26     & 0.54     & 0.0      & 0.43     & 0.43 \\
  DDC (Avg.)      & 0.73    & 0.68    & 0.9      & 0.69     & 0.59     & 0.03     & --       & -- \\ \hline
  RDDC            & \B{1.0} & \B{1.0} & 0.88     & 0.69     & \B{0.74} & \B{0.19} & \B{0.80} & \B{0.77} \\\thickhline
\end{tabular}
\egroup

\end{table}

The results of the experiments are listed in Tab. \ref{tab:results}. These show a large spread in performance between the different benchmark methods and between the different vectorization techniques. The highest performing vectorization technique also seems to be data-dependent, potentially making the choice difficult, especially if it has to be done in an unsupervised manner. On the Speech Commands dataset, all of the benchmark methods more or less fail, which indicates that they are unable to correctly model the temporal dependence in the data.

Recall that for the Twenty Newsgroups dataset, the Skipgram model already takes some of the temporal dependence into account by embedding nearby words close to each other. We conjecture that this is the cause for the increase in performance for the vector-based models, compared to the RNN-based model. The performance gap is especially visible for the time averaged vector representations.

\subsection{Qualitative analysis}
\begin{figure}[t] 
  \centering
  \begin{subfigure}{0.49\textwidth}
    \centering
    \includegraphics[width=0.8\textwidth]{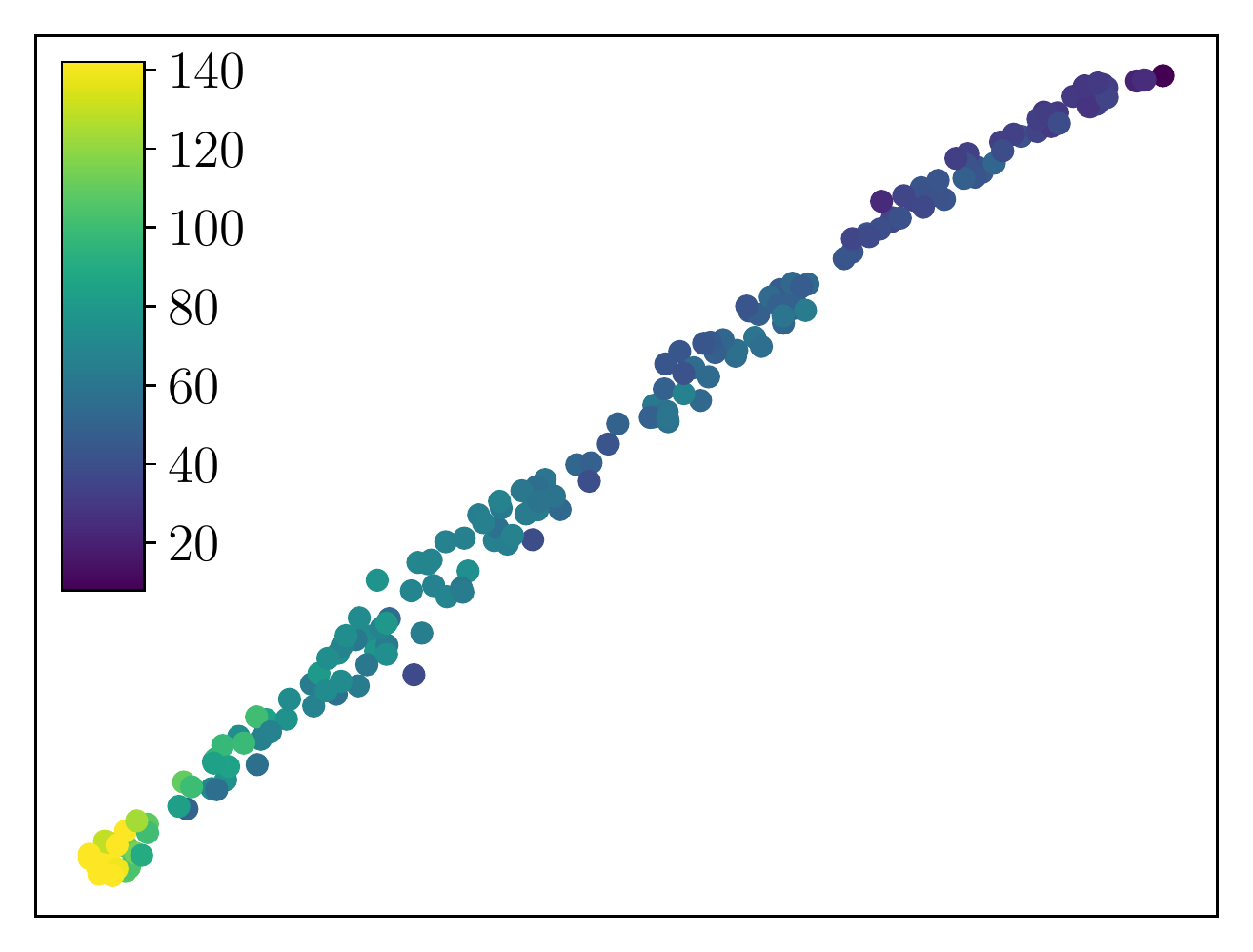}
    \caption{Sequence Length}
    \label{fig:tfspeech:seqlen}
  \end{subfigure}
  \begin{subfigure}{0.49\textwidth}
    \centering
    \includegraphics[width=0.8\textwidth]{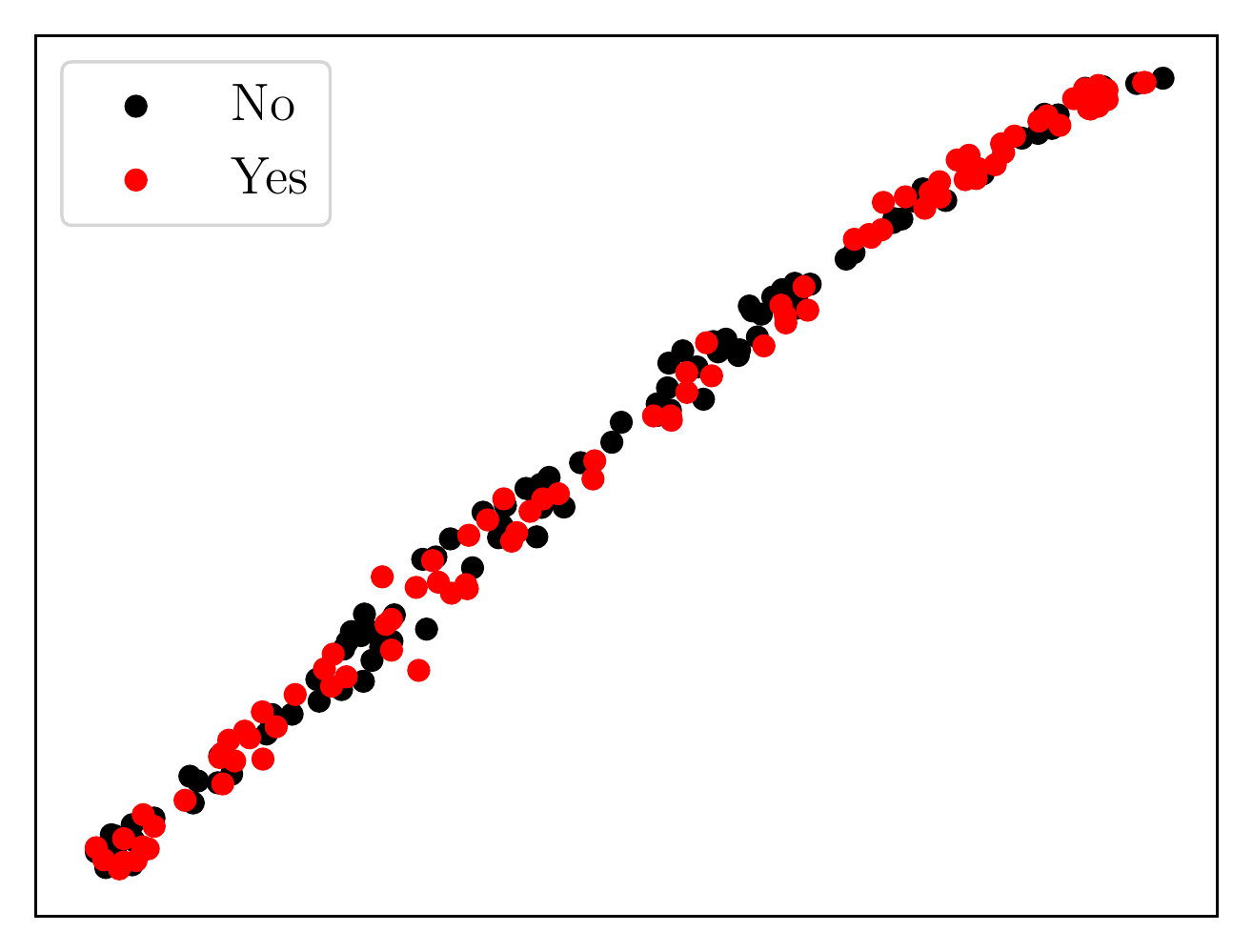}
    \caption{Ground truth}
    \label{fig:tfspeech:labels}
  \end{subfigure}

  \begin{subfigure}{0.49\textwidth}
    \centering
    \includegraphics[width=0.8\textwidth]{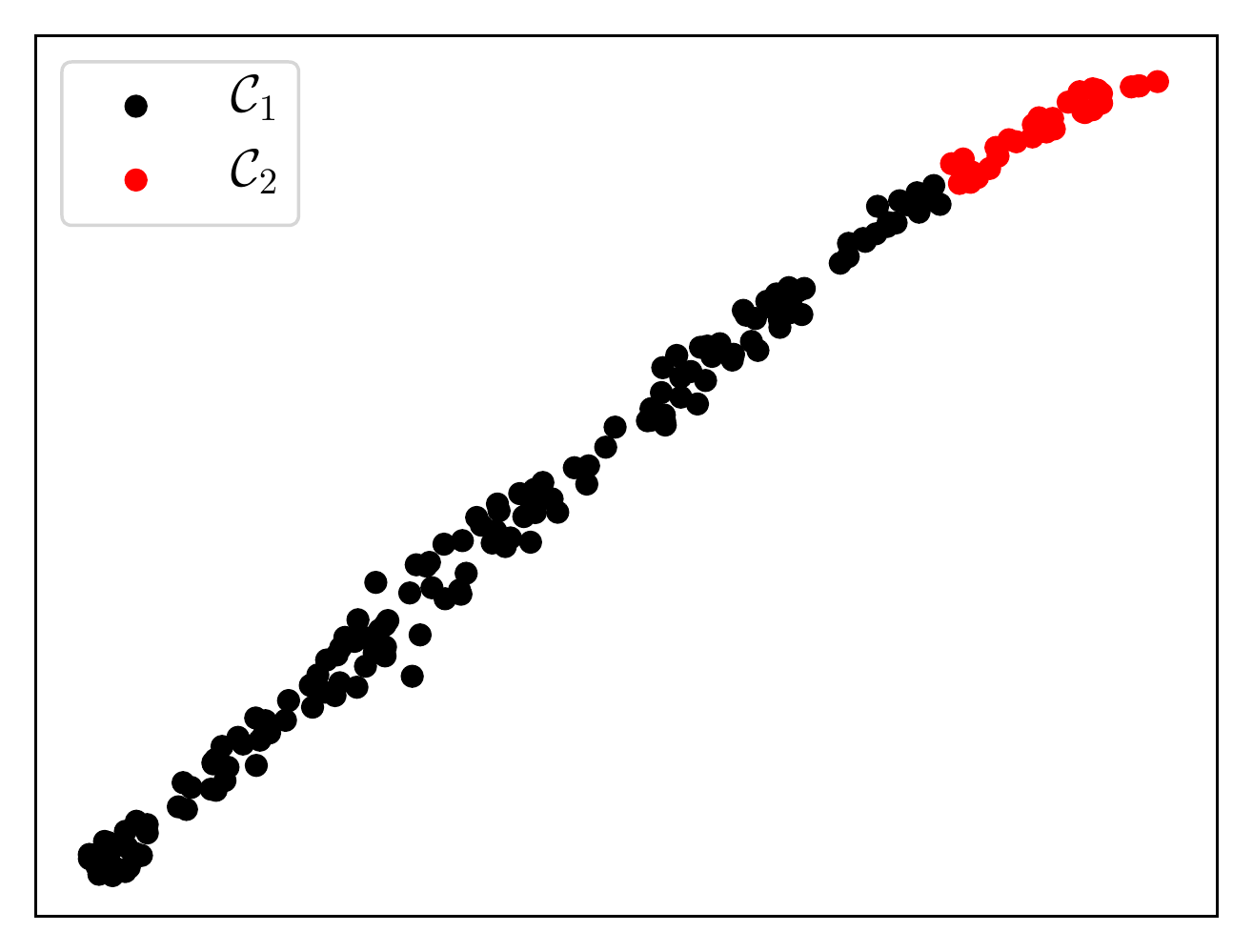}
    \caption{\( k \)-means predictions}
    \label{fig:tfspeech:kmpred}
  \end{subfigure}
  \begin{subfigure}{0.49\textwidth}
    \centering
    \includegraphics[width=0.8\textwidth]{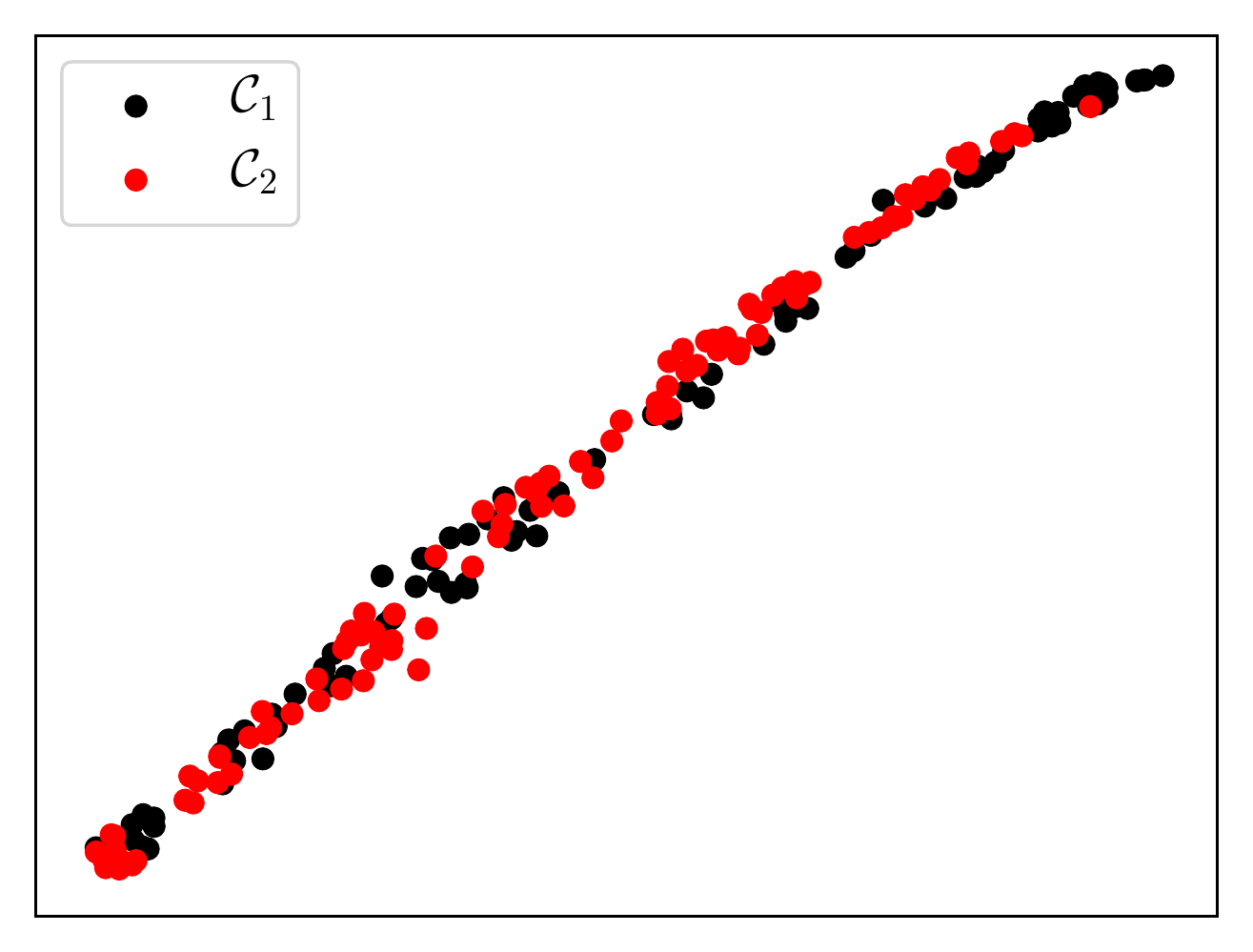}
    \caption{RDDC predicitions}
    \label{fig:tfspeech:rddcpred}
  \end{subfigure}
  \caption{\( t \)-SNE representation of time-averaged sequences from the Speech Commands dataset.}
  \label{fig:tfspeech}
\end{figure}

To further evaluate the validity of our results, we project the time-averaged Speech Commands data down to two dimensions using \( t \)-SNE \cite{van_der_maaten_visualizing_2008} (Fig. \ref{fig:tfspeech}). The points in Fig. \ref{fig:tfspeech:seqlen} indicate that the length of the sequences is a neighborhood determining feature. If we now consider the plot in Fig. \ref{fig:tfspeech:kmpred}, we see that \( k \)-means has learned to group sequences almost solely based on their lengths. In the event that sequence length was a reliable predictor for the class membership, this would be acceptable. However, this is not the case, as can be seen in Fig. \ref{fig:tfspeech:labels}. Shifting our focus to the predictions of RDDC (Fig. \ref{fig:tfspeech:rddcpred}), we see that RDDC instead learns features which are not directly related to the sequence lengths, making its predictions more accurate with respect to the ground truth labels.

\begin{figure}[t] 
  \centering
  \begin{subfigure}{0.49\textwidth}
    \centering
    \includegraphics[width=0.8\textwidth]{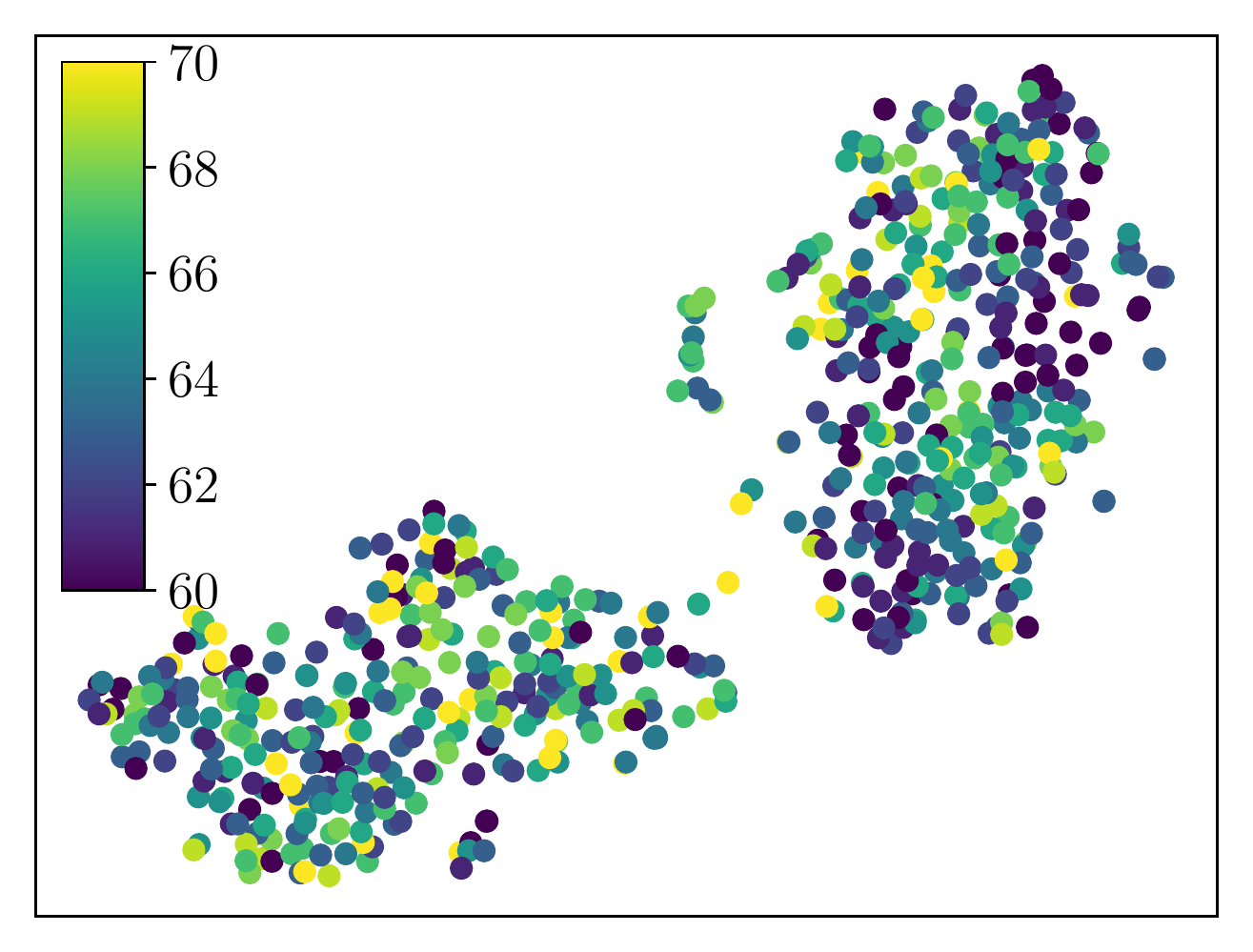}
    \caption{Sequence Length}
    \label{fig:tfspeech:seqlen_cropped}
  \end{subfigure}
  \begin{subfigure}{0.49\textwidth}
    \centering
    \includegraphics[width=0.8\textwidth]{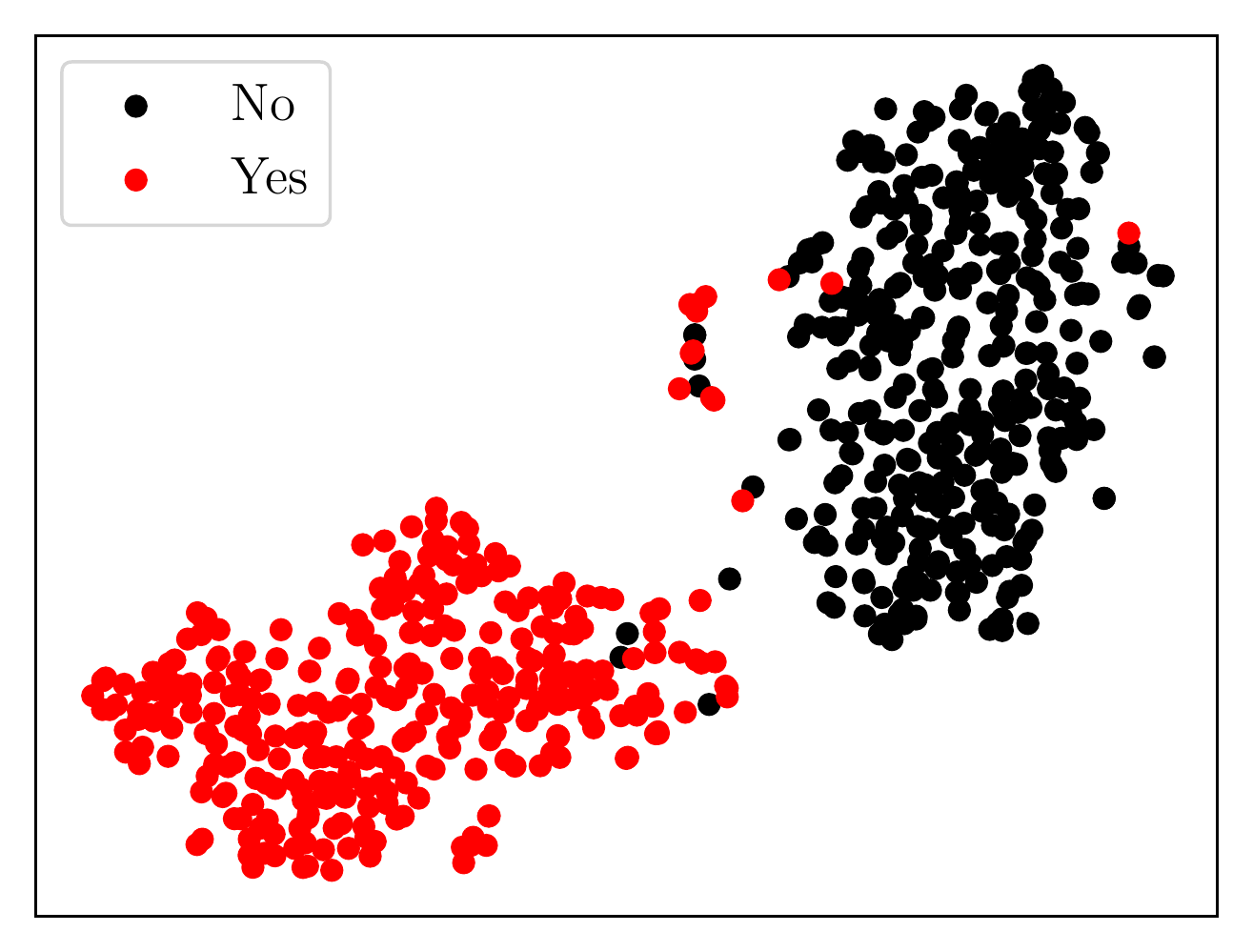}
    \caption{Ground truth}
    \label{fig:tfspeech:labels_cropped}
  \end{subfigure}

  \begin{subfigure}{0.49\textwidth}
    \centering
    \includegraphics[width=0.8\textwidth]{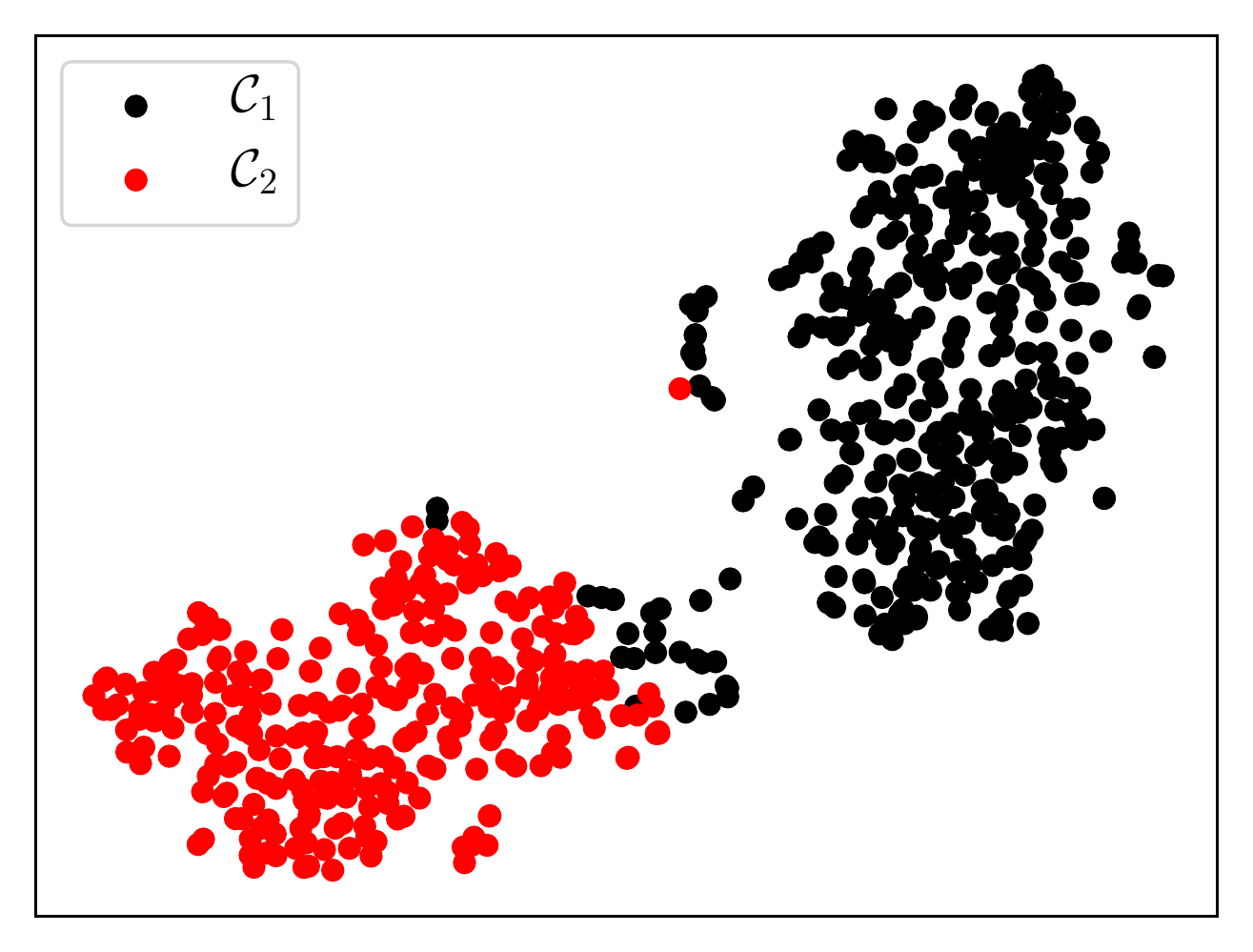}
    \caption{\( k \)-means predictions}
    \label{fig:tfspeech:kmpred_cropped}
  \end{subfigure}
  \begin{subfigure}{0.49\textwidth}
    \centering
    \includegraphics[width=0.8\textwidth]{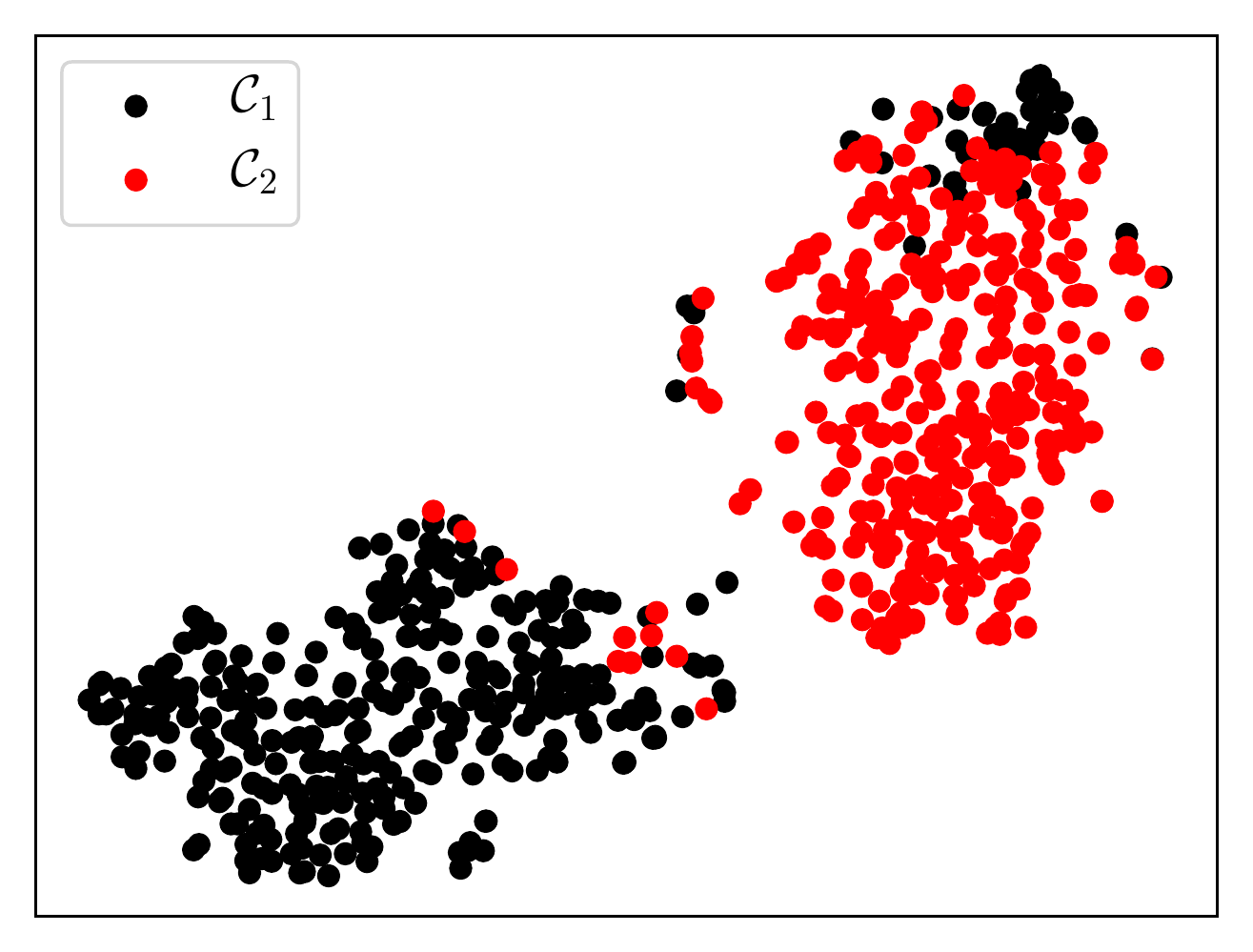}
    \caption{RDDC predicitions}
    \label{fig:tfspeech:rddcpred_cropped}
  \end{subfigure}
  \caption{\( t \)-SNE representation of sequences with lengths between \( 60 \) and \( 70 \) timesteps, from the Speech Commands dataset.}
  \label{fig:tfspeech_cropped}
\end{figure}

To eliminate the sequence length dependency, we remove all sequences shorter than \( 60 \) timesteps and longer than \( 70 \) timesteps. For \( t \)-SNE and \( k \)-means, the remaining sequences are then cropped to \( 60 \) timesteps, producing sequences of equal length. Fig. \ref{fig:tfspeech_cropped} shows the \( t \)-SNE representations of these sequences. From Fig. \ref{fig:tfspeech:seqlen_cropped}, it is indeed apparent that the sequence length dependency has been greatly reduced by considering sequences of approximately same length. Moreover, the \( t \)-SNE representation now shows two separate clusters, which correspond to the ground truth labels (Fig. \ref{fig:tfspeech:labels}).

Running \( k \)-means on only the sequences with similar lengths resulted in much improved predictions (Fig. \ref{fig:tfspeech:kmpred_cropped}), which was expected, due to the reduced influence of the sequence lengths. The RDDC predictions on the other hand, were obtained from the model trained on the full dataset. This further indicates that, for these sequences, RDDC trained on the full model has learned to separate the "Yes" and "No" recordings.

  \section{Conclusion}
    In this paper, we addressed the task of time series clustering. Our model uses a recurrent neural network as a feature extractor and a divergence-based clustering loss function in order to find underlying structure as well as optimize the feature extraction. Our approach is able to effectively cluster time series of different length and multivariate data with complex temporal dependencies, outperforming previous approaches that do not exploit the temporal dependencies in the data.

  \bibliography{bibl}
\end{document}